%% file: ebmc2026_main.tex
\documentclass[runningheads]{llncs}

\usepackage{eccv}

\usepackage{eccvabbrv}
\usepackage{graphicx}
\usepackage{booktabs}
\usepackage{soul}
\usepackage{algorithm, algorithmic}
\usepackage{multirow}

\usepackage[accsupp]{axessibility}  

\usepackage{hyperref}

\usepackage{orcidlink}

\begin{document}

\title{An Event Preserving Velocity Invariant Representation for Event Cameras} 

\titlerunning{An Event Preserving Velocity Invariant Representation for Event Cameras}

\author{Mikihiro Ikura\inst{1}\orcidlink{0000-0001-9258-3730} \and
Luna Gava\inst{1}\orcidlink{0000-0002-4240-8026}\and
Jiahang Wu\inst{1}\orcidlink{0009-0002-2794-3511}\and \\
Chiara Bartolozzi\inst{1}\orcidlink{1111-2222-3333-4444} \and
Arren Glover\inst{1}\orcidlink{0000-0003-3465-6449}}

\authorrunning{M.~Ikura et al.}

\institute{Istituto Italiano di Tecnologia, Via Morego 30 16163 Genova, Italy
\email{\{mikihiro.ikura,luna.gava,jiahang.wu,\\chiara.bartolozzi,arren.glover\}@iit.it}\\
}

\maketitle

\begin{abstract}
  Event cameras provide low-latency, high temporal resolution perception for real-time vision tasks such as robotics.The novel circuitry (i.e. asynchronous, independent pixels) that enables these advantages also introduces new algorithmic challenges. Velocity-invariant representations alleviate missing observations under slow motion and motion blur under fast motion, but most discard temporal information by converting events into image-like representations. We propose Set of Centre Active Receptive Fields (SCARF), a real-time velocity-invariant representation that preserves raw events while consistently handling fast motion, stationary scenes, and independently moving objects. SCARF achieves state-of-the-art performance in both computational efficiency and representation quality.
\end{abstract}

\section{Introduction}
\label{sec:intro}

Event cameras respond only to change in light and output a sparse, asynchronous stream of pixel activations. They produce low-latency and high dynamic range visual data with lower processing requirements compared to traditional cameras. Potential applications exists in applications that have power and processing constraints, such as mobile robotics, space, and wearable devices~\cite{gallego2020event}. As the data are sequentially produced and processed, rather than having the full frames of pixel readings at a single point in time, new algorithmic challenges must be addressed. Event representations are a common method to create pixel neighbourhoods that are fundamental to visual processing, and also allow downstream applications to be decoupled from the exact rate of data being produced~\cite{glover2021luvharris}.

\begin{figure}[t]
    \centering
    \includegraphics[width=0.8\linewidth]{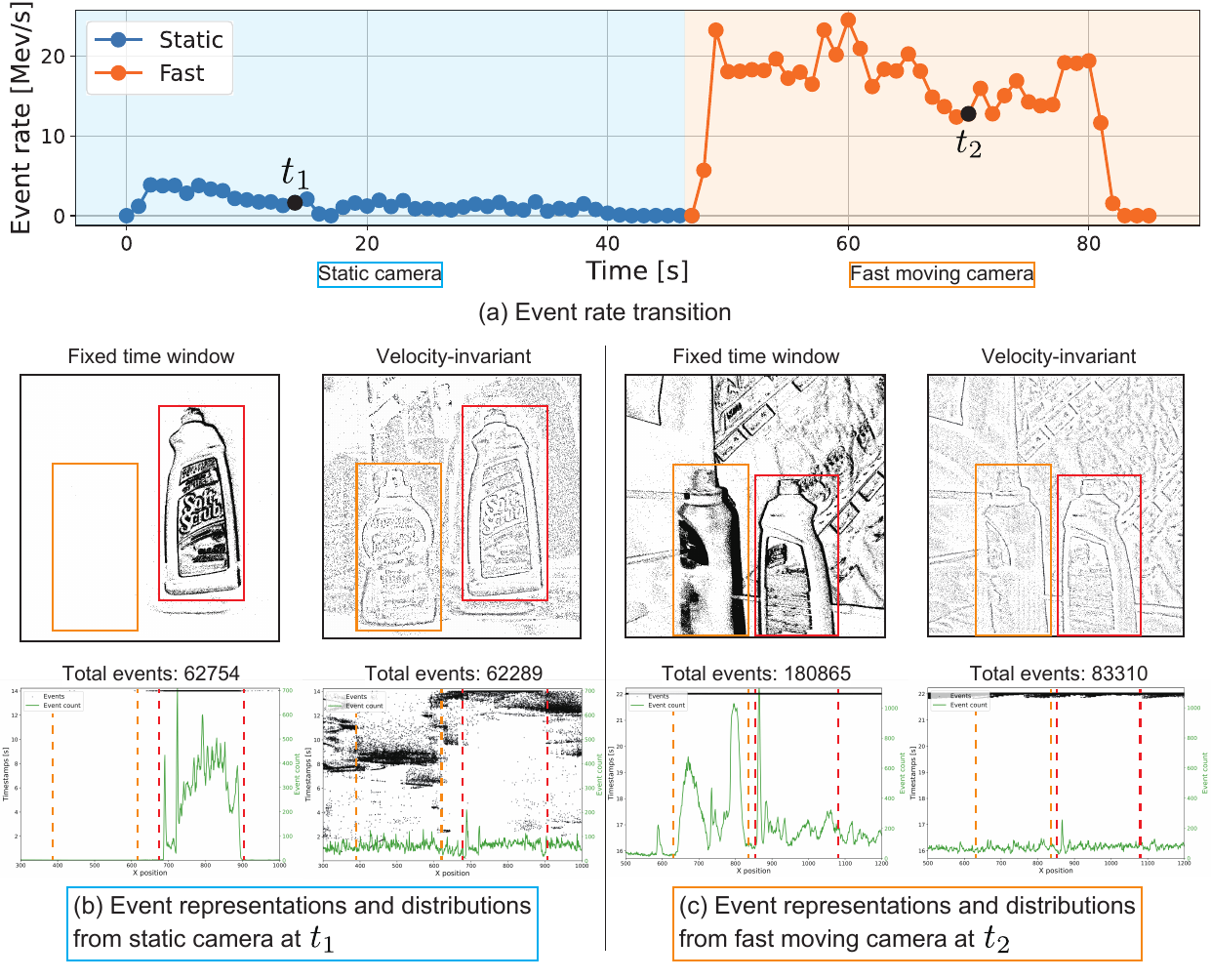}
    \caption{Velocity-independence of SCARF compared with a fixed temporal window (33 ms). Top: event-rate transition of the dataset, where two bottles and the event camera move independently. Middle: Fixed temporal window and SCARF. Bottom: event distributions (black) and event counts (green) along the x-axis. (a) The event rate varies significantly with scene motion. (b) The fixed temporal window fails to represent the slowly moving bottle due to insufficient events, whereas SCARF preserves a consistent representation by retaining older events. (c) For fast motion, the fixed temporal window introduces motion blur due to excessive events, while SCARF maintains sharp edges through consistent event storage.}
    \label{fig:velocity-independence}
\end{figure}

Event batches are a representation that capture the rich spatio-temporal information of the event stream that can then be encoded as a tensor~\cite{valerdi2023insights,scarpellini2021lifting,chen2022ecsnet,baldwin2022time} or directly processed~\cite{vasco2016fast,shiba2022event,mitrokhin2018event}. Motion-blur that occurs due to the temporal accumulation can be compensated for, e.g.~\cite{stoffregen2019event}, or implicitly within a network~\cite{zhu2018ev}. However batches assumes the relevant information occurs within a hard limit, typically 10-100 ms. In the wild, situations such as the camera stopping can cause algorithms to fail as data no longer contains the information expected (shown in Fig.~\ref{fig:velocity-independence}~(b)). On the other hand, exceedingly fast motion can also break the velocity consistency that is assumed algorithms (shown in Fig.~\ref{fig:velocity-independence}~(c)). Dynamic batch sizes cannot solve the common case in which there is more than one velocity profile - in this case an object that stops will still disappear from view. 

Velocity invariant representations~\cite{manderscheid2019speed,9095269,glover2021luvharris,9641205,annamalai2022event,10077556,10320049,glover2024edopt} can be used to more robustly represent scenes with various motions. Combined with event-by-event, continual updates of the representation, the temporal cut-off is removed such that fast, slow, and stopped objects are represented with consistency. Such representations are especially compatible when motion isn't a major feature of the task, such as feature~\cite{manderscheid2019speed,glover2021luvharris,10077556}, object~\cite{glover2024edopt}, or human pose~\cite{goyal2023moveenet} recognition, but most modify the visual data as some function of the raw events - resulting in a 2D array with arbitrary values. The raw events and timestamps are discarded, and any specific tensor encoding for a downstream task most likely cannot be achieved.

The proposed method, \underline{\textbf{S}}et of \underline{\textbf{C}}entre \underline{\textbf{A}}ctive \underline{\textbf{R}}eceptive \underline{\textbf{F}}ields (SCARF), is a representation with memory that achieves velocity consistency and also preserves raw events as well as timing information as shown in Fig.~\ref{fig:velocity-independence}~(b, c). SCARF is also lightweight, enabling real-time processing with high event resolution and event rates. The representation consists of dividing the sensor space into overlapping patches each with a circular buffer of a fixed amount of events. The intuition is that a fixed amount of texture is not valid for the full sensor array, but becomes more valid as the patch size decreases, and secondly that the overlapping region allow data to pass from patch to the next, as an object or edge passes across the scene. The paper describes the SCARF algorithm, compares to batch and velocity invariant representations, as well as performing an ablation study in regards to two downstream tasks: edge extraction and stereo depth estimation.

\section{Related Work}

Event cameras produce events that correspond to individual camera pixels changing their measured light level, and fire sparsely and asynchronously. Processing data in batches can lead to speed and performance improvements compared to individual event processing, but isn't robust to velocity variation, as illustrated in~\cref{fig:velocity-independence}. Event surfaces instead aim to integrate individual events into a data structure to achieve some useful benefits, such as event memory, maintaining a fixed size data representation, and processing over pixel neighbourhoods. The \textit{Time Surface (TS)} or \textit{Surface of Active Events (SAE)}~\cite{benosman2012asynchronous} explicitly represents timing information by storing a 2D array, $T$, with identical size as the camera sensor with the time of last firing $t$:
\begin{equation}
    T(x,y,p) \gets t
\end{equation}%
As the SAE achieves different representations for cameras or objects moving at different speeds, it is not \textit{velocity invariant}. Tasks such as object recognition and pose estimation are sensitive to velocity~\cite{iacono2018towards} and velocity invariance can make them more robust.

\begin{figure}[t]
    \centering
    \includegraphics[width=0.8\linewidth]{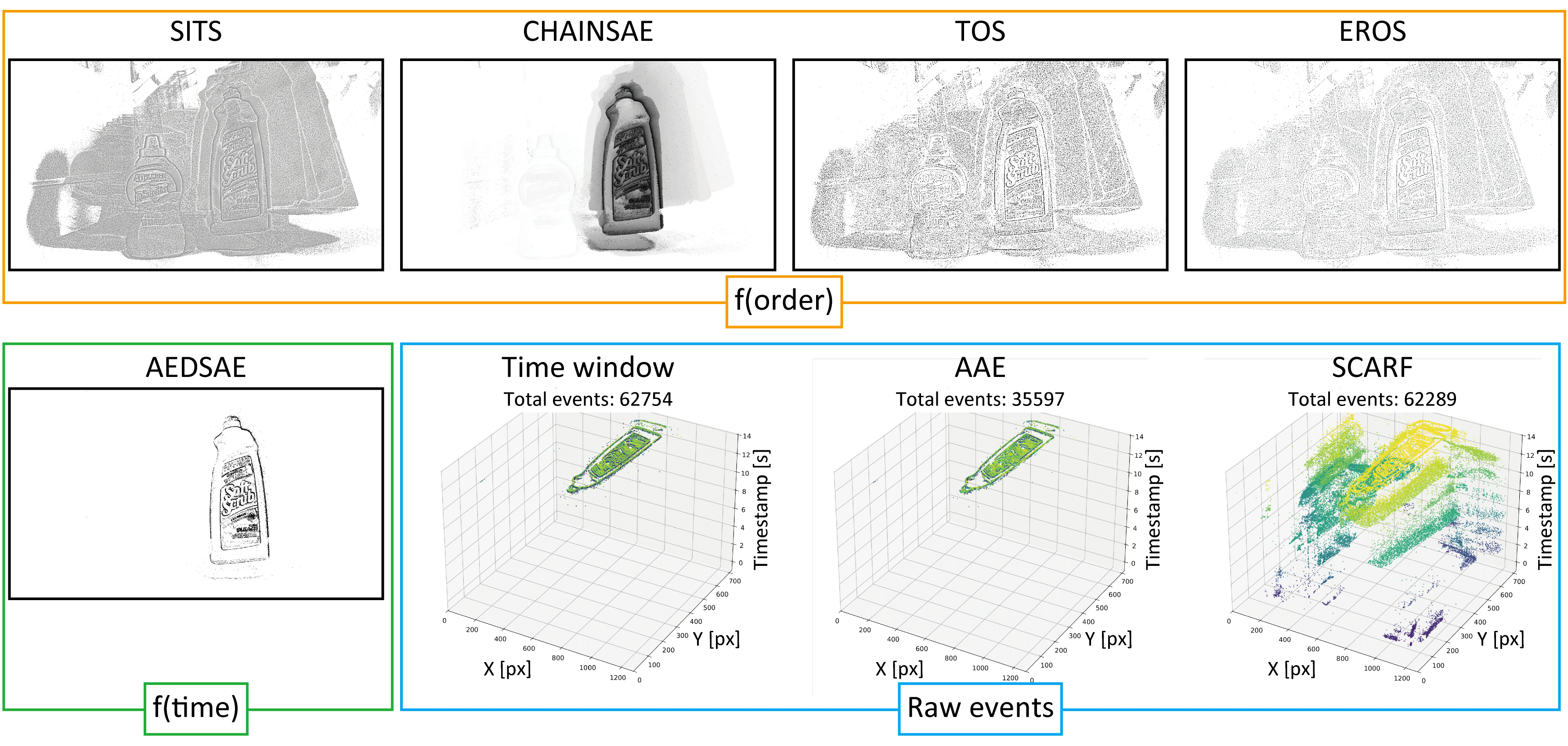}
    \caption{Output visualizations of state-of-the-art velocity-invariant event representations. SITS, CHAIN-SAE, TOS, and EROS construct 2D representations based on event order, while AED-SAE uses event timestamps. These methods discard the original timestamps and raw events in their final outputs. In contrast, AAE and SCARF preserve raw events while achieving velocity-invariant representations.}
    \label{fig:representations}
\end{figure}

``Event representation'' can refer to a wide range of topics including: how to encode data in a 2D/3D space for input into a neural network (e.g. VoxelGrid~\cite{zhu2019unsupervised} or TORE~\cite{baldwin2022time}) or representing the data with features (e.g. Hots~\cite{lagorce2016hots} or Hats~\cite{sironi2018hats}). These representations handle data encoding specifically for the network or downstream algorithm, but don't modify which data is used for the encoding. Typically a fixed size batch is used, which requires that the item of interest to have moved within the defined period. In this paper, we focus on algorithms which define which data should be encoded, specifically those that achieve velocity invariance. Many of these algorithms modify the data in some way, possibly making them incompatible with encoding methods, and propose that algorithms which both maintain raw events as well as achieving some invariance to velocity achieve the best of both worlds. We define the following characteristics for velocity invariant representations:
\begin{itemize}
    \item \textbf{Motion Blur}: typically all velocity invariant representations aim to remove motion blur.
    \item \textbf{Persistent/Time Limited}: some algorithms integrate all data incrementally, while others consider only a fixed length history. Time limited cannot achieve velocity invariance for stationary scenes.
    \item \textbf{Global/Local}: some algorithms consider a single velocity for the global scene, while others handle multiple different local velocities. 
    \item \textbf{Integration Metric}: the metric used to integrate data can generally come from a time parameter, a count of events, or recording the event order.
    \item \textbf{Output}: how is the data represented and what is available for downstream processing.
    \item \textbf{Event throughput}: how much processing is required per event.
    \item \textbf{Generation Speed}: how fast processing is required when querying the surface.
    \item \textbf{Accuracy}: how well the surface captures the features required for a downstream task.
\end{itemize}%
A summary and the output visualization of each algorithm according to these properties is shown in Table~\ref{tab:sota} and Fig.~\ref{fig:representations}, respectively, while an analysis of speed, throughput and accuracy forms the results of this paper. 

\begin{table}[t]
    \caption{A summary of state-of-the-art velocity invariant event-based representations.} \label{tab:sota}
    \centering
    {\scriptsize
    \begin{tabular}{|p{0.8cm}|p{2.3cm}|p{1.8cm}|p{1.6cm}|p{2.0cm}|p{2.2cm}|}
        \hline
        Year & Algorithm & Persistent/ Limited & Local/ Global & Underlying Integration & Output\\
        \hline
        2019 & SITS/  SILC~\cite{manderscheid2019speed}& Persistent  & Local & Order & pseudo-order \\
        2020 & CHAIN-SAE~\cite{9095269}& Persistent &  Global & Order & $f(\text{order})$ \\
        2021 & TOS~\cite{glover2021luvharris}& Persistent & Local & Order & $f(\text{pseudo-order})$\\
        2022 & event-LSTM~\cite{annamalai2022event} & Limited & Global & Time & $f(\text{time})$ \\
        2023 & AED-SAE~\cite{10077556}& Limited & Global & Time & $f(\text{time})$ \\ 
        2023 & VK-SITS~\cite{10320049}& Persistent & Local & Order & $f(\text{order})$ \\
        2023 & EROS~\cite{glover2024edopt}  & Persistent & Local & Order &  $f(\text{pseudo-order})$ \\
        2025 & AAE~\cite{niu2025esvo2}  & Limited & Local & Count & events \\
        \hline \hline
        2026 & SCARF (Ours) & Persistent & Local & Count & events \\ 
        \hline
    \end{tabular}
    }
\end{table}

The first velocity invariant representation was the Speed Invariant Temporal Surface (SITS)~\cite{manderscheid2019speed}, which records the pseudo-ordering of event arrival in local regions without requiring an expensive sorting operation over the entire sensor array. SITS introduced the idea of velocity invariance through event ordering as the same order of events theoretically occurs when moving quickly or slowly. The input is continuous, allowing long term memory, and local, allowing multiple simultaneous velocities. However, the algorithm performed pixel updates per event, reducing throughput, and also resulted in ``ghosting'' as targets moved.  

The Chain Surface of Active Events (CHAIN-SAE)~\cite{9095269} proposed to maintain the precise event order in a quickly updatable linked-list, with a precomputed non-linear function to reduce ghosting. The CHAIN-SAE was fast to compute, with long term memory, but due to its global nature could only robustly represent a single velocity. The parameters of the precomputed function should have also been modified based on the texture in the scene, making it effective in driving scenarios but unsuitable for irregularly motion of the camera or objects. While the authors proposed sub-regions CHAIN-SAE for smaller regions there existed no means to combine sub-regions, thereby requiring \textit{a-priori} knowledge of object size and location.

The Threshold Ordinal Surface (TOS)~\cite{glover2021luvharris} used local region updates similar to SITS and used a discrete cut-off function to reduce ghosting, but produced an almost binary image. To reduce artifacts introduced by the discrete function, the Exponentially Reduced Ordinal Surface (EROS)~\cite{glover2024edopt} instead used a smooth exponential decay function for 6-DoF object pose tracking, as well as human pose estimation~\cite{goyal2023moveenet}. Both of these representations convert the events to a 2D matrix structure that hold the filtered pseudo-order, and the raw events are no longer recoverable. They achieve long term memory and can represent multiple velocities simultaneously, but also have no method to remove noise in adverse conditions.

The Adaptive Exponential Decay Surface of Active Events (AED-SAE)~\cite{10077556} and the Accumulation of Events algorithm (AAE)~\cite{niu2025esvo2} both take a large batch of events and attempt to refine the set to remove motion blur. As the batch is of finite time, the methods only have short term memory and cannot represent stationary objects. AED-SAE estimates the number of edges in the scene using a frequency domain analysis, from which it dynamically adjusts an SAE normalisation parameter in time. The method is global, and best for a single velocity of the camera or an object. The AAE also uses a (different) statistical approach to decide the number of events in the final representation. However, it does so for each patch the scene is divided into and can represent different velocities in each patch. Additionally, as the events themselves are stored, they are also recoverable for sparse downstream processing.

Data driven methods for learning velocity invariant representations include the event-LSTM~\cite{annamalai2022event}, in which the features of the hidden layers of a temporal encoder-decoder aimed to become robust to velocity variation. A batch of data was required to be fed to the network resulting in limited temporal memory. The Variable Kernel Speed Invariant Surface (VK-SITS)~\cite{10320049} combined the velocity invariance of SITS with the learnability of TORE~\cite{baldwin2022time} to provide a modified time ordered feature for use as a neural network input.

To date, a comparison of these algorithms in terms of event-rate throughput, update rate, and performance on memory, motion-blur removal, and downstream tasks has not been performed.

\section{A Set of Centre Active Receptive Fields} \label{section:methods}

An event-camera produces an asynchronous stream of \textit{events} for each pixel, with a polarity (increase and decrease in light), and a timestamp, i.e. $<u,v,p,t>$. The event is only triggered if the light changes beyond a logarithmic threshold. In practice, the majority of events come from scene or object edges due to object motion, or the camera itself moving. Images are not extracted from the camera, rather individual pixel locations are asynchronously emitted from the sensor in the precise temporal order at which they occurred. An object moving quickly will produce more events than an identical slowly moving object, however for a significant number of vision algorithms it is often desirable that both objects are represented identically (velocity invariance) such that processing/analysis obtains an identical outcome.

We propose SCARF as a velocity invariant representation that is suitable for all conditions including fast and slow motion with multiple moving objects. Additionally, we focus on minimising event-throughput and allowing a high output frequency for suitability in real-time operation. The intuition behind SCARF is that, while a fixed amount of texture is not valid for the full sensor array, such an approach becomes more valid as the patch size decreases, and secondly that as an object moves across the sensor array, from one patch to the next, inactive regions are used to overwrite the data in active regions, allowing SCARF to remain sparse with areas of no texture.



The SCARF is composed of a set of local patches in which a fixed number of events, $N$, are stored in an ordered circular buffer. The full patch size is defined as $r.k \times r.k$ where an internal \textit{active} region of $k\times k$ is centered. The value of $r$ therefore defines the size ratio to an \textit{inactive} border around the central patch. The set of \textit{active} events will form the velocity invariant output of SCARF, while the \textit{inactive} border is necessary as it instigates the forgetting of no-longer-relevant events.

A patch has a fixed location on the sensor array such that events occurring in the overlaid region are added to the circular buffer, and the oldest event of the same buffer is removed. The event is also tagged as \textit{active} or \textit{inactive} depending on which region in the patch it occurs. The circular buffer will therefore be composed of a mix of \textit{active} and \textit{inactive} events. Only \textit{active} events contribute to the representation output, allowing SCARF to maintain sparsity if all data in a buffer is filled from \textit{inactive} regions. 

SCARF patches are tessellated across the sensor array such that the active regions of each patch align and cover the full array. Therefore, each pixel has a single \textit{active} region associated with it, but due to patch overlap of the \textit{inactive} region, can have several \textit{inactive} region associations. The representation used by downstream tasks is formed only from \textit{active} events across all patches. Active events can be extracted from each patch in two modes:
\begin{itemize}
    \item Sparse: extract the raw, sparse, events and iterate through the list for an algorithm, e.g. line detection, Hough transform.
    \item Dense: extract the optional 2D array, generated on-the-fly, which can be compatible with image-based vision processing, e.g. a convolutional neural network.
\end{itemize}

\begin{figure}[t]
    \centering
    \includegraphics[width=0.9\linewidth]{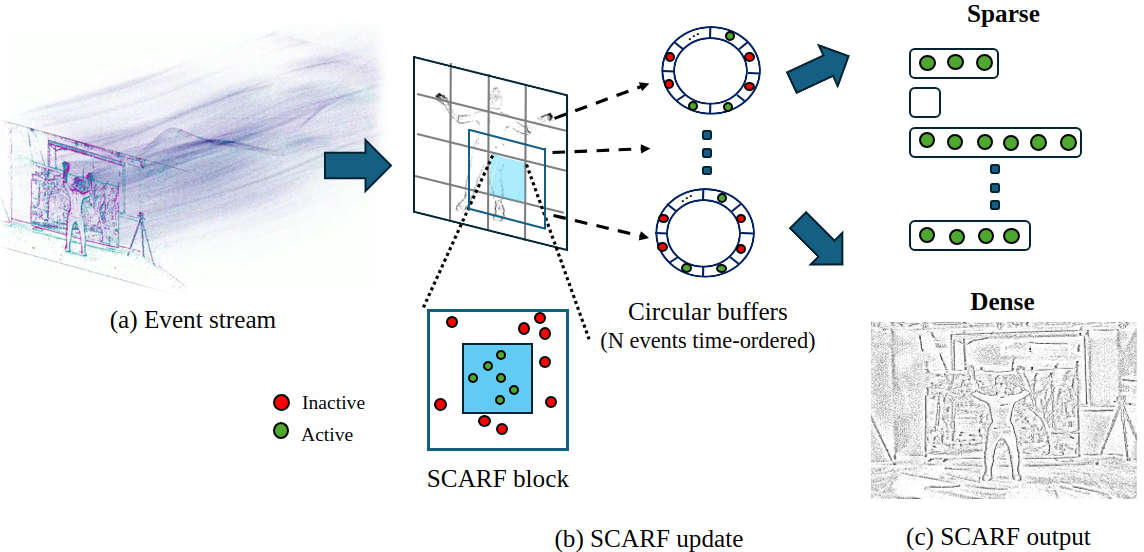}
    \caption{The SCARF pipeline: (a) event stream - asynchronous pixels with microsecond temporal resolution, (b) the set of overlapping blocks with active regions (blue) covering the entire sensor plane, (c) circular buffers store events in active (blue) and inactive (white) regions in temporal order, (d) visualising the contents of the buffers by drawing active events on the sensor plane.}
    \label{fig:introfigure}
\end{figure}

\subsubsection{Pixel-to-Patch Mapping}

SCARF is initialised to achieve a fast update by pre-computing the mapping between incoming pixel location and the associated patches. Each pixel in the sensor plane (of resolution $W \times H$) is covered by a patch in its \textit{active} region, and multiple other patches in their \textit{inactive} regions. SCARF is initialised with the set of patches $\{P\}$, a set of connection maps that relate each pixels in the sensor plane to receptive fields for all \textit{active} regions, $\{A\}$, and a similar set of connection maps for \textit{inactive} regions, $\{I\}$.

\subsubsection{Event-by-Event Update}

SCARF is updated for each event produced by the event-camera as it occurs. The data does not need to split into batches in any way. The connection maps are queried at each event's $<u,v>$ pixel position to identify associated $P$ and their corresponding circular buffers. The update procedure is simply to copy events into the circular buffers.  Events added to circular buffers from $\{A\}$ get tagged as active, $a=1$, and events added to circular buffers from connection maps $\{I\}$ get tagged as inactive, $a=0$.

The update procedure is cheap as it is a simple copy of data. The number of circular buffers that need to be updated scales with the ratio $r$, however, typically there are only 4 buffers per event. In contrast to other velocity invariant algorithms, there is no region update required, which means many less operations per event, and computational requirements do not scale with kernel size, $k$.


If image-like arrays are desired as an output of SCARF, it is also very cheap to iteratively construct such an image on-the-fly, event-by-event. When an event in the circular buffer is overwritten, it is first checked to see if it was an active event, and if so, the image at that pixel location is decremented by $c$. Each new active event is added to the image by increasing the pixel location by $c$. 

\begin{algorithm}[tb]
\footnotesize
\caption{SCARF Update (per event)}
\begin{algorithmic}[1]
\footnotesize
\REQUIRE $e = \langle u, v, p, t, a \rangle$ \COMMENT{\textit{new event with activity tag}}
\REQUIRE $B$ \COMMENT{\textit{circular buffer}}

\STATE $B(\{A\}_{uv}) \gets \langle u, v, p, t, a = 1 \rangle$ 
\COMMENT{\textit{mark active}}

\FOR{$i \in \{I\}_{uv}$}
    \STATE $B(i) \gets \langle u, v, p, t, a = 0 \rangle$
    \COMMENT{\textit{mark inactive}}
\ENDFOR

\end{algorithmic}
\label{alg:update}
\end{algorithm}

\subsection{Comparison with Velocity Invariant Representations}

SCARF is similar to SITS~\cite{manderscheid2019speed}, TOS~\cite{glover2021luvharris}, and EROS~\cite{glover2024edopt} as it has local updates to handle multiple velocities simultaneously and long-term persistence, however it doesn't require a kernel update for each event, resulting in a higher event throughput. Additionally, un-like VK-SITS~\cite{10320049}, SCARF can be queried for the raw sparse events, not only the dense image-like array. CHAIN-SAE~\cite{9095269} and AED-SAE~\cite{10077556} perform global updates making them more suited to a single velocity in the scene. AAE~\cite{niu2025esvo2} is similar to SCARF in that they are local and can generate sparse event outputs - however as AAE only operates over a limited memory, also like event-LSTM~\cite{annamalai2022event}, SCARF should achieve longer-term memory.

\section{Experiments and Results} \label{section:exp}
\renewcommand{\thefootnote}{\arabic{footnote}}\setcounter{footnote}{1}
Comparison is performed to velocity invariant representations by demonstrating particular failure cases, quantitatively evaluating robustness to motion blur, and measuring event throughput and algorithm update frequency. Finally two use cases are presented using the representations: depth estimation and line feature extraction. The SITS, TOS, CHAIN-SAE, EROS, AED-SAE, AAE and SCARF algorithms are re-implementated in the open source code\footnote[1]{https://github.com/robotology/event-driven} and compared. The main parameter of each algorithm is ablated using line feature extraction described in the supplementary as a metric. The VK-SITS and Event-LSTM implementations were not available and required training, so could not be compared validly. The temporal window was also compared as a baseline (non-velocity-invariant) representation. Encoding algorithms such as VoxelGrid~\cite{zhu2019unsupervised}, TORE~\cite{baldwin2022time}), Hots~\cite{lagorce2016hots} and Hats~\cite{sironi2018hats} are not valid for comparison as they don't modify the batch used and therefore belong to the temporal window evaluation. The advantage of SCARF, as it doesn't modify the underlying data and keeps raw events, is that the output of SCARF could be encoded with such an algorithm to achieve a velocity invariant coding with e.g. VoxelGrid/TORE.

\subsection{Persistence and Multi-object Speeds}

\begin{figure*}[t]
    \centering
    \includegraphics[width=0.9\linewidth]{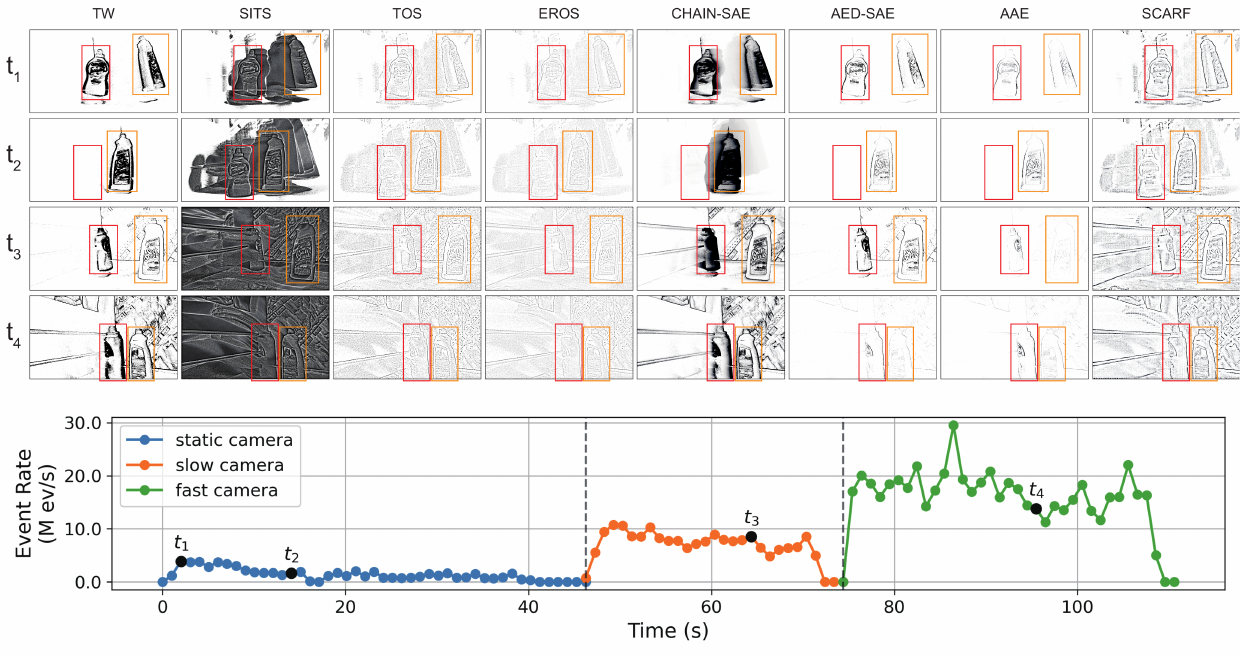}
    \caption{Qualitative comparison of event-based representations during various scenarios: ($t_1$) static camera, two moving objects, ($t_2$) static camera, one moving object, ($t_3$) slow-moving camera, one moving object, ($t_4$) fast-moving camera, one moving object.}
    \label{fig:qualitative_multi-objects-speeds}
\end{figure*}

To particularly highlight issues that can arise ``in-the-wild'' for autonomous perception systems, but which are not always present in typical driving or object moving datasets, a qualitative comparison was performed with a self-recorded dataset which contains camera motion and motion of multiple objects. Specifically, there are periods of non-motion (which commonly occurs) of both objects and cameras, as shown in Fig.~\ref{fig:qualitative_multi-objects-speeds}. As a baseline it can be seen that the Time Window (TW) consistently exhibited motion blur and/or fading because it was tuned for a single temporal window (i.e. velocity) and therefore cannot accommodate different object or camera speeds simultaneously.
CHAIN-SAE produced consistent representation for camera motion but motion blur when objects moved independently, e.g. $t_2$ and $t_3$, additionally the non-moving object was lost when the secondary object was moved. TW, AAE, and AED-SAE also lost the stationary object during secondary motion at $t_2$ as they have a limited temporal memory. SITS, TOS, EROS, and SCARF maintained visual signal of both objects under all conditions, however have differences in their representation values. SITS had a ghosting effect due to its decay design, while TOS and EROS had reasonable noise artefacts ``left behind'' in non-moving parts of the scene. SCARF also had noise artefacts, but possibly achieved a stronger contrast between true edges and noise.

Motion blur and fading can effect downstream algorithms which assume a consistent visual input over time. Noise also effects downstream algorithms, but if uniformly distributed, can be more easily disregarded by algorithms looking for strong features for vision tasks. The local and persistent properties of SCARF (and others) are important for operating ``in-the-wild'' in which consistent object and camera motion cannot be guaranteed.

\subsection{Motion Blur}

\begin{figure}[t] 
\centering 
\centering 
\includegraphics[width=0.8\linewidth]{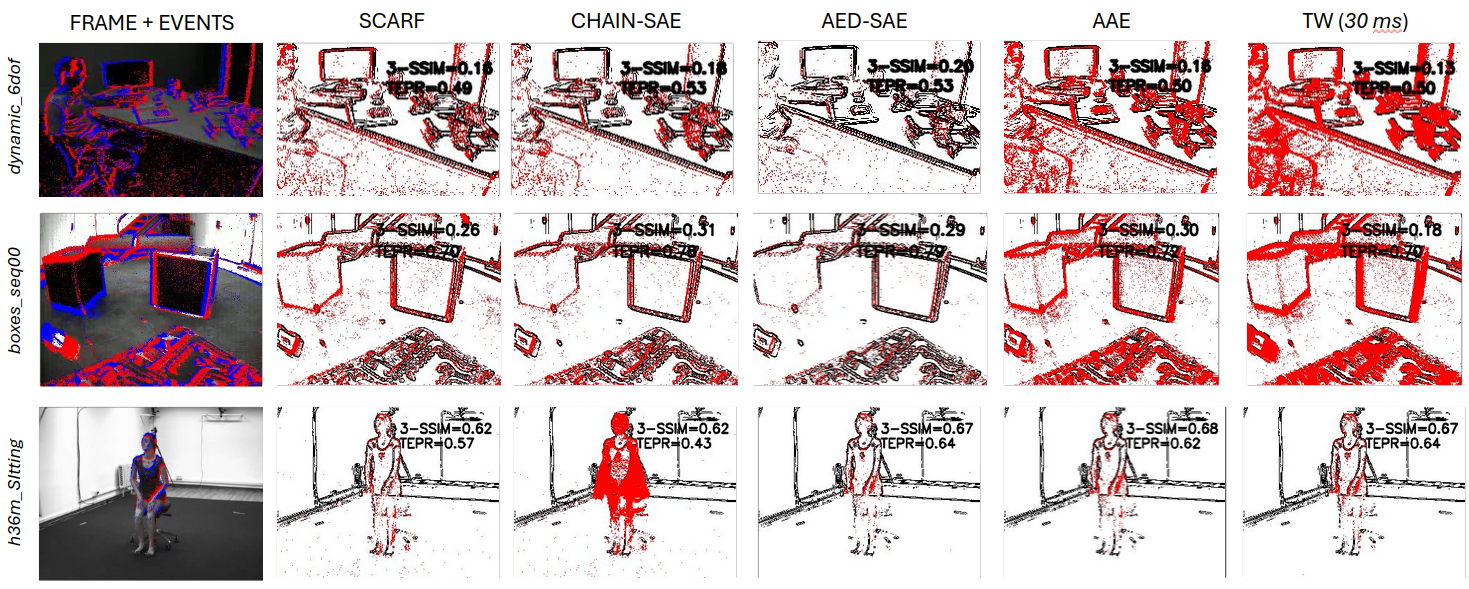} 
\caption{Comparison of some event representation for sequences from different datasets. First column shows grayscale image and events with polarities. Other columns show the events of the representation (red), the Sobel edges (black) extracted from the grayscale image. 3-SSIM and TEPR scores are provided.} \label{fig:datasets_overlay} \end{figure}

To quantify the velocity invariant output in representing edges, while removing motion blur, the algorithms were compared across several publicly available sequences, see Fig.~\ref{fig:datasets_overlay}. The available sequences were limited and chosen to give fairness to all algorithms, and therefore contained typically only a single velocity of motion, and from datasets with RGB aligned frames. ECD \textit{dynamic\_6dof}~\cite{mueggler2017event}, EVIMO \textit{boxes\_seq00}~\cite{mitrokhin2019ev}, and event-H36m \textit{h36m\_Sitting}~\cite{goyal2023moveenet} were therefore chosen. The RGB frames served as a proxy ground truth by extracting Sobel-filtered greyscale images to identify pixel locations in which events should be located in the velocity invariant representation. As in \cite{niu2025esvo2}, $3$-SSIM and TEPR metrics are used to compare velocity invariant representation and the Sobel images (computed only on pixels where event data is present). 

Results on the three representative sequences in \cref{fig:datasets_overlay} are reported in~\cref{tab:dyn_shapes_poster}. SCARF achieved performance comparable to the other representations, although not consistently the highest scores. While methods such as AAE and CHAIN-SAE often obtain higher metric values, qualitative overlays (Fig.~\ref{fig:velocity-independence}) reveal that they produce thicker edges or motion blur. In contrast, SCARF generates consistently thin, sharp contours that better reflect velocity invariance, but are penalized by $3$-SSIM and TEPR, which inherently favour denser edge responses. This suggests that current metrics may not fully capture the structural fidelity and thin-edge characteristics that SCARF is designed to preserve.

\begin{table*}[t]
\centering
\scriptsize
\caption{Mean$\pm$std of 3-SSIM and TEPR for sequences from three different datasets.}
\label{tab:dyn_shapes_poster}
\setlength{\tabcolsep}{3pt}
\renewcommand{\arraystretch}{1.1}
\begin{tabular}{l|cc|cc|cc}
\toprule
& \multicolumn{2}{c|}{dynamic\_6dof} & \multicolumn{2}{c|}{boxes\_seq00} & \multicolumn{2}{c|}{h36m\_Sitting}\\
Reps & 3-SSIM & TEPR & 3-SSIM & TEPR & 3-SSIM & TEPR \\
\midrule
scarf     & 0.28$\pm$0.11 & 0.46$\pm$0.15 & 0.26$\pm$0.03 & \textbf{0.68}$\pm$0.04 & 0.61$\pm$0.03 & 0.52$\pm$0.04 \\
eros      & 0.18$\pm$0.09 & 0.38$\pm$0.14 &  0.11$\pm$0.02 & 0.36$\pm$0.05 & 0.57$\pm$0.04 & 0.40$\pm$0.06 \\
sits      & 0.07$\pm$0.05 & 0.31$\pm$0.10 & 0.08$\pm$0.02 & 0.32$\pm$0.05 & 0.56$\pm$0.04 & 0.43$\pm$0.04 \\
tos       & 0.14$\pm$0.09 & 0.38$\pm$0.13 & 0.08$\pm$0.03 & 0.36$\pm$0.05 & 0.57$\pm$0.04 & 0.42$\pm$0.05 \\
chainsae  & 0.30$\pm$0.11 & \textbf{0.50}$\pm$0.16 & \textbf{0.31}$\pm$0.04 & 0.64$\pm$0.08 & 0.61$\pm$0.02 & 0.35$\pm$0.08  \\
aedsae    & \textbf{0.33}$\pm$0.09 & 0.51$\pm$0.17 & 0.27$\pm$0.05 & 0.53$\pm$0.05 & 0.66$\pm$0.005 & 0.64$\pm$0.05\\
aae       & 0.28$\pm$0.13 & 0.49$\pm$0.14 &  0.25$\pm$0.04 & 0.50$\pm$0.05 & \textbf{0.67}$\pm$0.004 & \textbf{0.65}$\pm$0.07 \\
tw (30 ms) & 0.27$\pm$0.10 & 0.50$\pm$0.14 & 0.15$\pm$0.03 & 0.63$\pm$0.06 &  0.66$\pm$0.004 & 0.63$\pm$0.06  \\
\bottomrule
\end{tabular}
\end{table*}

\subsection{Algorithm Runtime}
\begin{figure}[t]
\centering

\begin{minipage}{0.5\linewidth}
    \centering
    \includegraphics[width=0.9\linewidth]{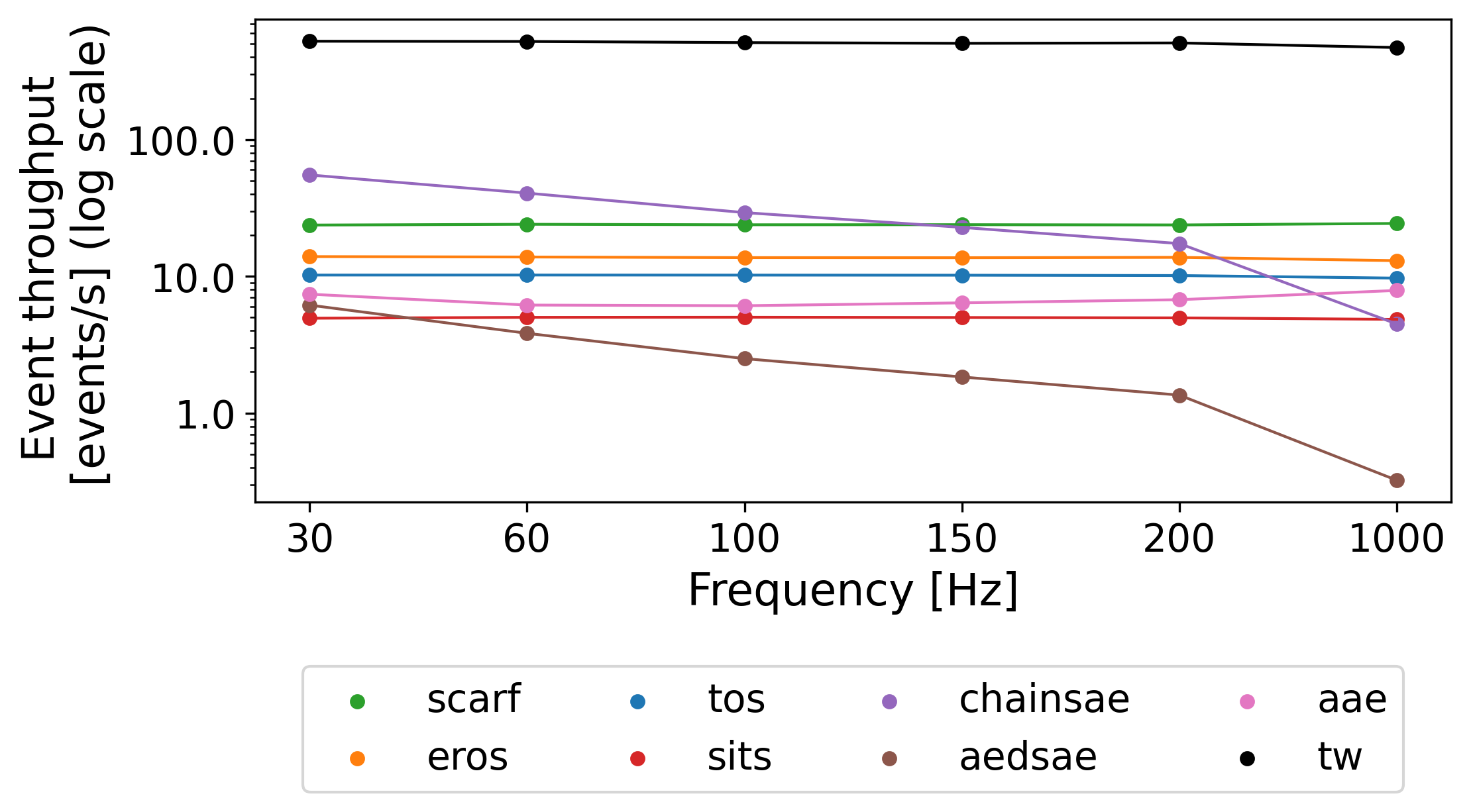}
\end{minipage}
\hfill
\begin{minipage}{0.38\linewidth}
    \scriptsize
    \centering
    \begin{tabular}{lcc}
    \hline
    \textbf{Rep.} & \textbf{C/E} [$\mu s$] & \textbf{C/U} [$\mu s$]\\
    \hline
    SCARF & 0.070 & 0.057 \\
    SITS & 0.217 & 8.646 \\
    TOS & 0.096 & 7.324 \\
    EROS  & 0.084 & 6.940 \\
    CHAIN-SAE & 0.026 & 218.488 \\
    AED-SAE & 0.073 & 2786.166 \\
    AAE & 0.033 & 250.767 \\
    TW & 0.064 & 0.060 \\
    \hline
    \end{tabular}
\end{minipage}

\caption{Comparison of event throughput (left) and computational cost (right) per event (C/E) and per update (C/U) computed using \textit{dynamic\_6dof}.}
\label{fig:throughput_cost}
\end{figure}

The computational performance of the representations was further analysed by measuring event throughput as a function of the surface update per event (C/E) and surface query frequency (C/U), i.e., how often a downstream algorithm requires the representation (30–1000 Hz). The distinction is made as some algorithms front-load computation for each event, while others backload computation only when the representation is required. The \textit{dynamic\_6dof} sequence was used.
As shown in~\cref{fig:throughput_cost}, the baseline Time Window (TW) maintains nearly constant throughput around 500 M events/s, since surface updates only require setting the corresponding pixel value e.g., to 255, and retrieving the representation requires no additional computation. 
Velocity invariant representation CHAIN-SAE could handle more than 20 M events/s (C/E), but, as computation was back-loaded, presented a steep throughput drop beyond 200 Hz due to its computational cost on retrieval (C/U), as illustrated in Fig.~\ref{fig:throughput_cost}. SCARF instead maintained constant run-time with increasing frequency, with a low C/E cost, allowing it to sustain at least 20 M events/s across almost all tested frequencies. EROS, TOS, and SITS also achieved a constant throughput at 13 M, 10 M and 5 M events/s respectively. AAE demonstrated some fluctuation with C/U frequency but was consistently below TOS. Finally, AED-SAE followed a similar trend to CHAIN-SAE due to algorithm back-loading, with significant degradation at higher refresh rates. Overall, for high frequency applications using an event camera SCARF achieves the highest throughput velocity invariant representation, while for low frequency tasks CHAIN-SAE would be more suitable.

\subsection{Depth Estimation}

\begin{figure}[t]
    \centering
    \begin{subfigure}[b]{\textwidth} \centering
    \includegraphics[width=0.9\linewidth]{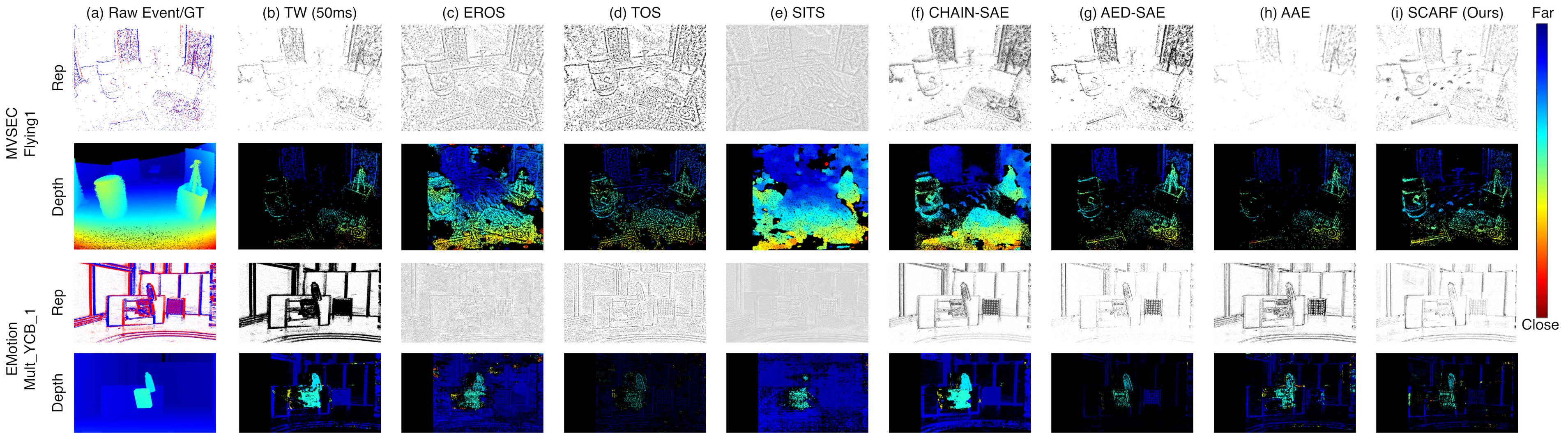}
    \caption{} \label{fig:depth_result}
    
    \end{subfigure}

    \begin{subfigure}[b]{0.4\textwidth} \centering
        \includegraphics[width=\linewidth]{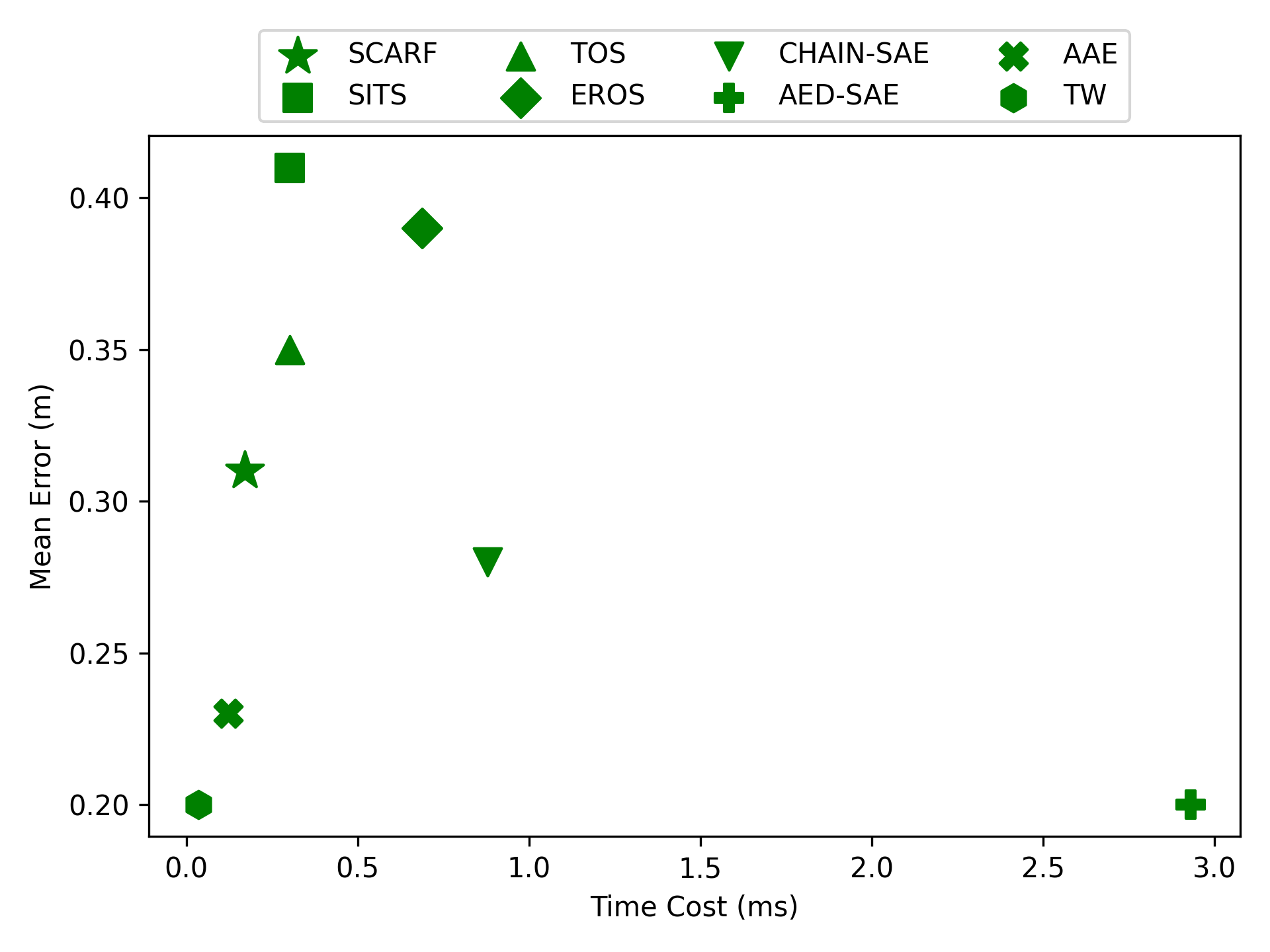}
         \caption{} \label{fig:mvsec_mevstc}
    \end{subfigure}%
    \begin{subfigure}[b]{0.4\textwidth} \centering
        \includegraphics[width=\linewidth]{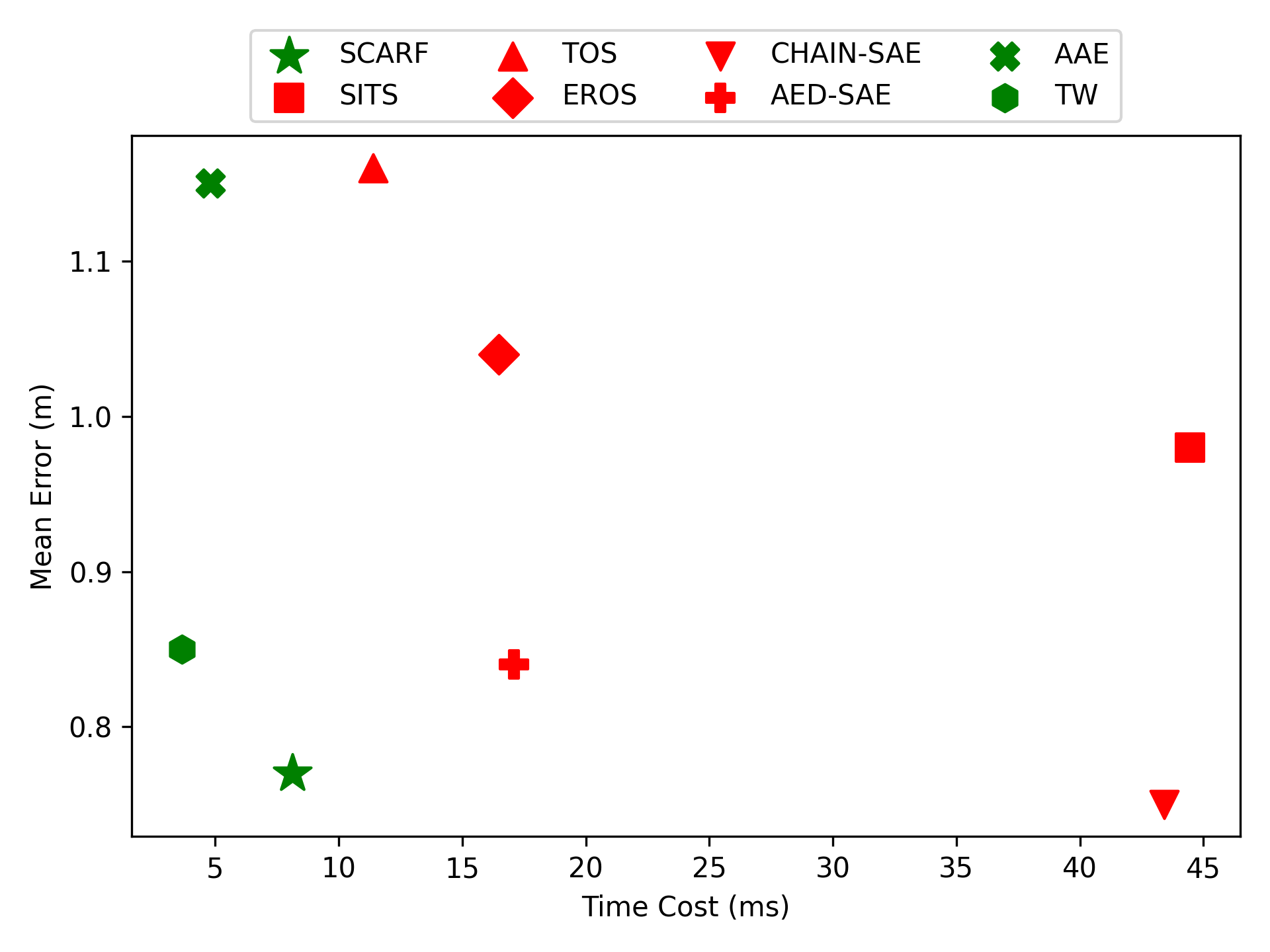}
        \caption{}\label{fig:emotion_mevstc}
    \end{subfigure}
    
    \caption{Depth estimation results and Time cost analysis. a) Qualitative evaluation of depth estimation with different event representations. Depth mean error versus time cost per frame at 100Hz depth output is shown for the b) MVSEC (Flying1, 346$\times$260, 0.20 Mev/s) and c) EMotion (Mult\_YCB\_1, 1280$\times$720, 19.61 Mev/s) datasets. Green indicates methods that achieved real-time performance at 100 Hz given the dataset event rate, while red indicates non-real-time operation.}
    \label{fig:depth}
\end{figure}

To compare event representations on real tasks, we evaluate the algorithm performance for stereo depth estimation, using Semi-Global Block Matching (SGBM)~\cite{hirschmuller2007stereo} as a back-end. The SGBM algorithm was designed for frame-based images, but was adapted to event representations by applying a Gaussian smoothing to the representations and adjusting SGBM parameters. Finally the output depth map is masked to produce the sparse depth map. The MVSEC~\cite{zhu2018multivehicle} dataset and a self-recorded dataset, in order to demonstrate on HD ($1280\times720$ pixels) stereo cameras, were used. 

The main noticeable qualitative difference in~\cref{fig:depth_result} is the density. While high density (see EROS, SITS, CHAIN-SAE) is traditionally preferred, the velocity representation should theoretically only represent edge locations; excess artefacts in the scene don't necessarily correspond to correct positions and this is reflected in higher error of these algorithms, as seen in~\cref{fig:mvsec_mevstc} and~\cref{fig:emotion_mevstc}. TOS, AED-SAE, AAE, and SCARF all produce sparse outputs with a variety of accuracies, but with AED-SAE having significantly higher processing rate. Of the remaining algorithm AAE achieves the best accuracy on MVSEC in~\cref{fig:mvsec_mevstc} and SCARF achieves the better accuracy on the high-resolution self-recorded dataset in~\cref{fig:emotion_mevstc}. As AAE was designed with the assumption of driving datasets the outcome is not unexpected. We note that the TW achieves a strong performance in both cases as the parameters were tuned for each dataset. Finally, for low resolution cameras all algorithms can run in real-time, while for the high resolution cameras only the TW, AAE, and SCARF can operate in real-time.

\subsection{Line Segment Feature Extraction}

\begin{figure}[t]
    \centering
    \begin{subfigure}[b]{\textwidth} \centering
    \includegraphics[width=0.9\linewidth]{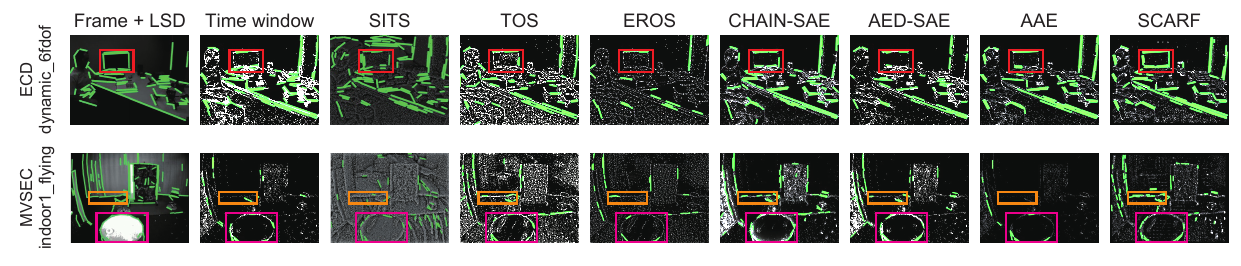}
    \caption{} \label{fig:line_qual}
    \end{subfigure}
    \begin{subfigure}[b]{0.4\textwidth} \centering
    \includegraphics[width=\linewidth]{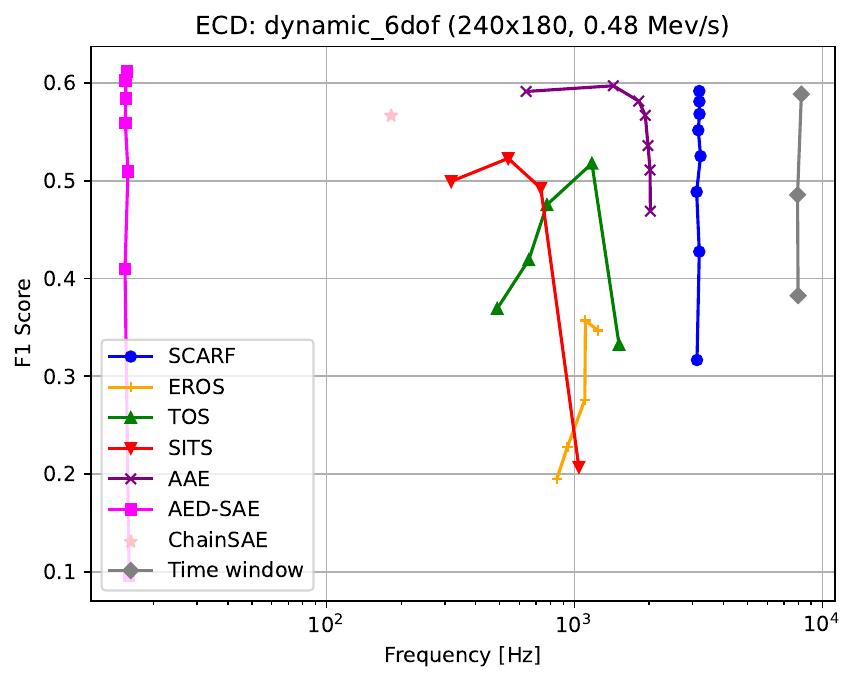}
    \caption{} \label{fig:line_ECD}
    \end{subfigure}%
    \begin{subfigure}[b]{0.4\textwidth} \centering
    \includegraphics[width=\linewidth]{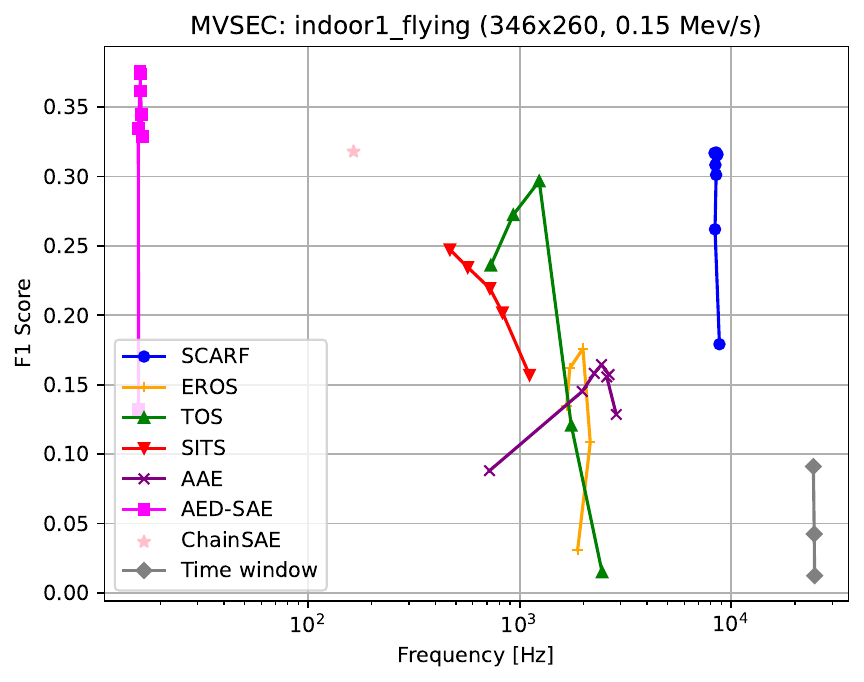}
    \caption{} \label{fig:line_MVSEC}
    \end{subfigure}
    
    \caption{Results of line segment extraction from event representations. (a) Qualitative comparison of line segments extracted by LSD, with boxes highlighting representative regions. (b, c) F1 score versus processing frequency (representation output fixed at 100 Hz; log-scale x-axis) on ECD ``\textit{dynamic\_6dof}'' and MVSEC ``\textit{indoor1\_flying}'', respectively.}
    \label{fig:f1_vs_freq}
\end{figure}

As a second downstream task, we evaluated frame-based line segment extraction using the Line Segment Detector (LSD)~\cite{Grompone2012}. Accuracy was measured using line heat maps~\cite{Zhou2019, Huang2018} on ECD ``\textit{dynamic\_6dof}'' (240x180) and MVSEC ``\textit{indoor1\_flying}'' (346x260), with a tolerance of 1\% of the image diagonal~\cite{Huang2018}. F1 scores were computed at 30 Hz over 10 s. To evaluate the trade-off between accuracy and efficiency, we measured processing frequency while fixing the representation output rate to 100 Hz.


\cref{fig:line_ECD} and~\cref{fig:line_MVSEC} compare F1 score and processing frequency for different parameter settings described in the supplementary. The Time window achieved the highest processing frequency but required dataset-specific tuning, resulting in inconsistent accuracy across different event rates. EROS, TOS, SITS, and AAE exhibit a clear accuracy--speed trade-off. Although AED-SAE achieved the highest F1 scores, it was also the slowest method. CHAIN-SAE produced accuracy comparable to SCARF but at a lower processing frequency. Overall, SCARF provides the most favorable balance between accuracy and efficiency, making it well suited to real-time robotic applications.

\section{Discussion} \label{section:discuss}

While we propose SCARF as a velocity invariant representation suitable for all conditions, the condition for long term persistence comes also with the unavoidable occurrence of also integrating and remembering noise, which is visible in many of the qualitative results presented. As perfect conditions never exist, it is possible to minimise the problem by adjusting camera sensitivity and focus correctly before experimentation, or perhaps new generations of camera will be developed with autonomous focus and sensitivity control. For already acquired dataset we have anecdotally noticed that a speckle/salt-and-pepper filter can lead to a cleaner SCARF output. As a counter point it has also been noted that good performance can be achieved even with noisy-looking SCARF output. In such a case, it is possible that the noise is typically uniform and downstream algorithms can easily ignore it as there are no strong features present.

Real-time operation is also an interesting discussion point. In the present datasets, SCARF and others were able to achieve real-time operation, however, datasets exist that produce a higher event rate such that all algorithms will fail to operate in real-time. As such thought needs to be made in tuning event camera sensitivity before recording and the decision for lower resolution cameras can be a suitable choice for certain operating conditions.




\section{Conclusion} \label{section:conc}



We evaluated all available velocity-invariant event representations and showed that each is suited to different tasks, motion profiles, and computational constraints. We introduced SCARF as a general-purpose representation designed to handle diverse scenarios, including missing objects, motion blur, depth estimation, and line extraction, while maintaining high event throughput. Across all evaluations, SCARF achieved competitive or superior performance, providing a strong balance between robustness and real-time efficiency.

Beyond existing datasets, many practical edge cases, such as stationary objects that cease generating events, remain under-represented. We therefore expect velocity-invariant representations to play an important role as a robust input layer for real-world autonomous systems, including robots and wearable devices.

\bibliographystyle{splncs04}
\bibliography{main.bib}
\include{ebmc2026_supplementary_asinclude}
\end{document}

%% file: ebmc2026_supplementary_asinclude.tex

\title{An Event Preserving Velocity Invariant Representation for Event Cameras \\
— Supplementary Material —} 

\titlerunning{An Event Preserving Velocity Invariant Representation for Event Cameras}

\author{Mikihiro Ikura\inst{1}\orcidlink{0000-0001-9258-3730} \and
Luna Gava\inst{1}\orcidlink{0000-0002-4240-8026}\and
Jiahang Wu\inst{1}\orcidlink{0009-0002-2794-3511}\and \\
Chiara Bartolozzi\inst{1}\orcidlink{1111-2222-3333-4444} \and
Arren Glover\inst{1}\orcidlink{0000-0003-3465-6449}}

\authorrunning{M.~Ikura et al.}

\institute{Istituto Italiano di Tecnologia, Via Morego 30 16163 Genova, Italy
\email{\{mikihiro.ikura,luna.gava,jiahang.wu,\\chiara.bartolozzi,arren.glover\}@iit.it}\\
}

\maketitle

\section{Algorithms}

\cref{alg:init} defines the pixel-to-patch mapping calculated once when initialising SCARF, such that a fast event throughput can be achieved.

\begin{algorithm}[t]
\caption{SCARF Pixel-to-Patch Mapping}
\begin{algorithmic}[1]
\footnotesize
\REQUIRE $\{P\}_{ij}$ \COMMENT{set of patches indexed in 2D}
\REQUIRE $\{A\}_{uv}$  \COMMENT{map: pixels $\rightarrow$ P}
\REQUIRE $\{I\}_{uv}$  \COMMENT{map: pixels $\rightarrow$ P}

\FOR{$u,v \in W,H$}
    \STATE $\{A\}_{uv} \gets \{P\}_{\frac{u}{k},\,\frac{v}{k}}$
\ENDFOR

\FOR{$i,j \in \frac{W}{k},\,\frac{H}{k}$}
    \FOR{$k(i-r) < u < k(i+r+1)$}
        \FOR{$k(j-r) < v < k(j+r+1)$}
            \IF{$\{P\}_{ij} \notin \{A\}_{uv}$}
                \STATE $\{I\}_{uv} \gets \{P\}_{ij}$
            \ENDIF
        \ENDFOR
    \ENDFOR
\ENDFOR

\end{algorithmic}
\label{alg:init}
\end{algorithm}

\section{Parameters}

SCARF is defined by the parameters defined in Table~\ref{tab:parameters}.

\begin{table}[t]
\caption{Parameter descriptions and default values required by SCARF.} \label{tab:parameters}
    \centering
    \small{
    \begin{tabular}{p{0.1\columnwidth}|p{0.6\columnwidth}|p{0.15\columnwidth}}
        \hline
        Name & Description & Default\\
        \hline
        $k$ & pixel height and width of the active region  & 9 \\
        $r$ & ratio of in-active to active region & 1.2 \\
        $\alpha$ & linear tuning parameter for $N$ & 2.0 \\
        $N$ & number of events in the circular buffer & $\alpha(k \cdot r)^2$ \\
        $c$ & pixel intensity accumulation per event & 0.3 \\
        \hline
    \end{tabular}
    }
\end{table}

Furthermore, Table~\ref{tab:X_parameters} shows parameters used to visualize the trade-off between accuracy and process frequency in the downstream task experiment of line segment feature extraction.
\begin{table}[t]
\caption{Hyper-parameters from each baseline for the evaluation of line segment features. All bold values with underlines are default values and other values are used for F1 score distributions of line segment accuracy.}
\centering
\scalebox{0.8}{
\begin{tabular}{@{}llc@{}}
    \toprule
    \textbf{Baselines} & \multicolumn{2}{c}{\textbf{Hyper-parameters}} \\
    \midrule
    Time window & window size [ms] & 10, 20, \underline{\textbf{33}} (30 Hz) \\ \midrule
    SITS & kernel size & 5, 7, \underline{\textbf{9}}, 13 \\ \midrule
    \multirow{2}{*}{%
      \begin{tabular}[c]{@{}l@{}}
        TOS
      \end{tabular}%
    } & kernel size & 3, 5, \underline{\textbf{9}}, 13, 17 \\
     & decay & \underline{\textbf{2.0}} \\ \midrule
    \multirow{2}{*}{%
      \begin{tabular}[c]{@{}l@{}}
        EROS
      \end{tabular}%
    } & kernel size & 3, 5, \underline{\textbf{9}}, 13, 17 \\
     & decay & \underline{\textbf{0.3}} \\ \midrule
    CHAIN-SAE & \multicolumn{2}{c}{None} \\ \midrule
    \multirow{2}{*}{%
      \begin{tabular}[c]{@{}l@{}}
        AED-SAE
      \end{tabular}%
    } & lambda & 0.1, \underline{\textbf{0.3}}, 0.5, 0.7, 0.9, 1.1, 1.3 \\
    & delta T & \underline{\textbf{0.3}} \\ 
    \midrule
    AAE & block size [px] & 3, 7, 11, \underline{\textbf{15}}, 19, 23, 27 \\ \midrule
    SCARF & $\alpha$ & 1.0, 1.5, \underline{\textbf{2.0}}, 2.5, 3.0, 3.5, 4.0, 4.5 \\
    \bottomrule
    \end{tabular}
    }
    \label{tab:X_parameters}
\end{table}

\section{Results}

\subsection{Motion Blur}
Results on the other sequences from the same datasets of the main document are reported in~\cref{tab:dyn_shapes_poster,tab:boxes_tabletop_floor,tab:h36m}.

\begin{table*}[t]
\centering
\scriptsize
\caption{Mean$\pm$std of 3-SSIM and TEPR for ECD Shapes, and Poster datasets.}
\label{tab:dyn_shapes_poster}
\setlength{\tabcolsep}{3pt}
\renewcommand{\arraystretch}{1.1}
\begin{tabular}{l|cc|cc|}
\toprule
& \multicolumn{2}{c|}{shapes\_6dof} & \multicolumn{2}{c|}{poster\_6dof}\\
Reps & 3-SSIM & TEPR & 3-SSIM & TEPR  \\
\midrule
scarf     & 0.53$\pm$0.07 & 0.57$\pm$0.24 & 0.10$\pm$0.05 & 0.45$\pm$0.08 \\
eros      & 0.21$\pm$0.10 & 0.20$\pm$0.15 & 0.08$\pm$0.05 & 0.40$\pm$0.07 \\
sits      & 0.07$\pm$0.08 & 0.12$\pm$0.08 & 0.04$\pm$0.03 & 0.39$\pm$0.06\\
tos       & 0.14$\pm$0.11 & 0.21$\pm$0.16 & 0.05$\pm$0.04 & 0.40$\pm$0.07 \\
chainsae  & 0.50$\pm$0.04 & 0.54$\pm$0.16 & 0.11$\pm$0.06 & \textbf{0.51}$\pm$0.10 \\
aedsae    & 0.54$\pm$0.06 & 0.54$\pm$0.27 & 0.11$\pm$0.04 & 0.50$\pm$0.11\\
aae       & \textbf{0.54}$\pm$0.08 & 0.64$\pm$0.21 &  \textbf{0.12}$\pm$0.06 & 0.47$\pm$0.08 \\
tw (30 ms) & 0.51$\pm$0.08 & \textbf{0.68}$\pm$0.19  &  0.07$\pm$0.05 & 0.48$\pm$0.08\\
\bottomrule
\end{tabular}
\end{table*}

\begin{table*}[t]
\centering
\scriptsize
\caption{Mean$\pm$std of 3-SSIM and TEPR for EVIMO Table-top, and Floor datasets.}
\label{tab:boxes_tabletop_floor}
\setlength{\tabcolsep}{3pt}
\renewcommand{\arraystretch}{1.1}
\begin{tabular}{l|cc|cc|}
\toprule
& \multicolumn{2}{c|}{table\_top\_seq00} & \multicolumn{2}{c|}{floor\_seq00} \\
Reps & 3-SSIM & TEPR & 3-SSIM & TEPR \\
\midrule
scarf     & \textbf{0.17}$\pm$0.03 & \textbf{0.53}$\pm$0.06 & 0.20$\pm$0.01 & 0.73$\pm$0.02 \\
eros      & 0.08$\pm$0.03 & 0.38$\pm$0.05 & 0.13$\pm$0.01 & 0.60$\pm$0.01 \\
sits      & 0.06$\pm$0.03 & 0.37$\pm$0.05 & 0.12$\pm$0.01 & 0.62$\pm$0.03 \\
tos       & 0.06$\pm$0.03 & 0.39$\pm$0.05 & 0.14$\pm$0.02 & 0.62$\pm$0.02 \\
chainsae  & \textbf{0.17}$\pm$0.03 & 0.52$\pm$0.08 & \textbf{0.26}$\pm$0.02 & \textbf{0.81}$\pm$0.02 \\
aedsae    & 0.15$\pm$0.03 & 0.50$\pm$0.08 & 0.20$\pm$0.02 & \textbf{0.81}$\pm$0.02 \\
aae       & \textbf{0.17}$\pm$0.03 & 0.51$\pm$0.06 & 0.21$\pm$0.03 & 0.80$\pm$0.02 \\
tw (30 ms) & 0.10$\pm$0.03 & 0.49$\pm$0.07 & 0.21$\pm$0.03 & 0.80$\pm$0.02 \\
\bottomrule
\end{tabular}
\end{table*}

\begin{table*}[t]
\centering
\scriptsize
\caption{Mean$\pm$std of 3-SSIM and TEPR for event-H36m Posing datasets.}
\label{tab:h36m}
\setlength{\tabcolsep}{3pt}
\renewcommand{\arraystretch}{1.1}
\begin{tabular}{l|cc|}
\toprule
& \multicolumn{2}{c|}{h36m\_Posing} \\
Reps & 3-SSIM & TEPR\\
\midrule
scarf     & 0.42$\pm$0.08 & 0.37$\pm$0.11 \\
eros      & 0.33$\pm$0.10 & 0.23$\pm$0.08 \\
sits      & 0.32$\pm$0.10 & 0.25$\pm$0.08 \\
tos       & 0.32$\pm$0.10 & 0.24$\pm$0.09 \\
chainsae  & 0.49$\pm$0.07 & 0.31$\pm$0.04 \\
aedsae    & 0.64$\pm$0.01 & 0.66$\pm$0.06 \\
aae       & \textbf{0.65}$\pm$0.007 & \textbf{0.70}$\pm$0.08 \\ 
tw & 0.64$\pm$0.01 & 0.68$\pm$0.07 \\ 
\bottomrule
\end{tabular}
\end{table*}

\subsection{Depth}
Precise numbers about results on depth estimation for different representations are reported in ~\cref{tab:depth}

\begin{table*}[htbp]
    \scriptsize
    \centering
    \caption{Quantitative evaluation of depth estimation with different event representations.}
    \setlength{\tabcolsep}{3pt}
    \begin{tabular}{l l c c c c}

        \toprule
        \multicolumn{1}{l}{Data Sequence}
            & \multicolumn{1}{l}{Rep.} 
            & \multicolumn{1}{c}{Mean Err$\pm$(Std)} 
            & \multicolumn{1}{c}{Median Err}
            & \multicolumn{1}{c}{Relative Err}
            & \multicolumn{1}{c}{\#Pixels}  \\
            
        \multicolumn{1}{l}{}
            & \multicolumn{1}{l}{} 
            & \multicolumn{1}{c}{[m] $\downarrow$} 
            & \multicolumn{1}{c}{[m] $\downarrow$}
            & \multicolumn{1}{c}{[m] $\downarrow$}
            & \multicolumn{1}{c}{[million]} \\
            
        \midrule
        \textit{MVSEC} & TW (50ms) &0.21$\pm$(0.36) & 0.09 & 9\% & 0.55\\
        \textit{(flying1)} & TW (100ms) & 0.26$\pm$(0.41) & 0.12 & 10\% & 0.08\\
        & EROS & 0.39$\pm$(0.57) & 0.18 & 14\% & 2.86\\
        & TOS & 0.35$\pm$(0.52) & 0.16 & 13\% & 0.73\\
        & SITS & 0.41$\pm$(0.57) & 0.19 & 15\% & 4.77\\
        & CHAIN-SAE & 0.28$\pm$(0.44) & 0.13 & 11\% & 2.91\\
        & AED-SAE & \textbf{0.20}$\pm$(\textbf{0.35}) & \textbf{0.10} & \textbf{8\%} & 0.84\\
        & AAE & 0.23$\pm$(0.38) & 0.11 & 10\% & 0.52\\
        & SCARF (Ours) & 0.31$\pm$(0.48) & 0.14 & 12\% & 0.66\\

        \midrule
        \textit{E-Motion} & TW (50ms) & 0.85$\pm$(1.44) & 0.26 & 13\% & 2.76 \\
        \textit{mult\_ycb\_1} & TW (100ms) & 1.04$\pm$(1.60) & 0.35 & 16\% & 1.68\\
        & EROS & 0.90$\pm$(1.36) & 0.33 & 14\% & 3.73 \\
        & TOS & 1.16$\pm$(1.84) & 0.33 & 17\% & 0.93 \\
        & SITS & 0.98$\pm$(1.36) & 0.43 & 17\% & 4.47 \\
        & CHAIN-SAE &  \textbf{0.75}$\pm$(1.31) & \textbf{0.23} & \textbf{11\%} & 3.24 \\
        & AED-SAE & 0.84$\pm$(1.56) & \textbf{0.23} & 12\% & 0.38 \\
        & AAE & 1.15$\pm$(1.73) & 0.40 & 16\% & 1.49 \\
        & SCARF (Ours) &  0.77$\pm$(\textbf{1.29}) & 0.25 & \textbf{11\%} & 0.76 \\
        
        \bottomrule
    \end{tabular}
    \label{tab:depth}
\end{table*}

\subsection{Video}
We provide a video showing the comparison of event-based representations for publicly available datasets and our recorded dataset. First, we display the matching between sobel-edges (black) and the representation events (red) to show motion-blur of some representations on the three sequences described in the main document. Secondly, we show the different representations on three sequences of the in-house recorded dataset, where the objects stop or slow-down their motion. Finally, we show results on downstream tasks: depth estimation and line detection.


%% file: main.bib
@String(ICCV= {Int. Conf. Comput. Vis.})

@String(ICCV  = {ICCV})

@article{annamalai2022event,
  title={Event-LSTM: An unsupervised and asynchronous learning-based representation for event-based data},
  author={Annamalai, Lakshmi and Ramanathan, Vignesh and Thakur, Chetan Singh},
  journal={IEEE Robotics and Automation Letters},
  volume={7},
  number={2},
  pages={4678--4685},
  year={2022},
  publisher={IEEE}
}

@inproceedings{manderscheid2019speed,
  title={Speed invariant time surface for learning to detect corner points with event-based cameras},
  author={Manderscheid, Jacques and Sironi, Amos and Bourdis, Nicolas and Migliore, Davide and Lepetit, Vincent},
  booktitle={Proceedings of the IEEE/CVF Conference on Computer Vision and Pattern Recognition},
  pages={10245--10254},
  year={2019}
}

@ARTICLE{10077556,
  author={Li, Jinjian and Su, Li and Guo, Chuandong and Wang, Xiangyu and Hu, Quan},
  journal={IEEE Sensors Journal}, 
  title={Asynchronous Event-Based Corner Detection Using Adaptive Time Threshold}, 
  year={2023},
  volume={23},
  number={9},
  pages={9512-9522},
  doi={10.1109/JSEN.2023.3257329}}

@ARTICLE{9095269,
  author={Lin, Shijie and Xu, Fang and Wang, Xuhong and Yang, Wen and Yu, Lei},
  journal={IEEE Robotics and Automation Letters}, 
  title={Efficient Spatial-Temporal Normalization of SAE Representation for Event Camera}, 
  year={2020},
  volume={5},
  number={3},
  pages={4265-4272},
  doi={10.1109/LRA.2020.2995332}}

@INPROCEEDINGS{10320049,
  author={Acin, Laure and Jacob, Pierre and Simon-Chane, Camille and Histace, Aymeric},
  booktitle={2023 Twelfth International Conference on Image Processing Theory, Tools and Applications (IPTA)}, 
  title={VK-SITS: a Robust Time-Surface for Fast Event-Based Recognition}, 
  year={2023},
  volume={},
  number={},
  pages={1-6},
  doi={10.1109/IPTA59101.2023.10320049}}

@INPROCEEDINGS{9641205,
  author={Hu, Rui and Xia, Yuanqing and Sun, Zhongqi},
  booktitle={2021 IEEE International Conference on Unmanned Systems (ICUS)}, 
  title={A Less Noisy Time Surface for Event-based Visual Odometry}, 
  year={2021},
  volume={},
  number={},
  pages={299-304},
  doi={10.1109/ICUS52573.2021.9641205}}

@article{glover2021luvharris,
  title={luvharris: A practical corner detector for event-cameras},
  author={Glover, Arren and Dinale, Aiko and Rosa, Leandro De Souza and Bamford, Simeon and Bartolozzi, Chiara},
  journal={IEEE Transactions on Pattern Analysis and Machine Intelligence},
  volume={44},
  number={12},
  pages={10087--10098},
  year={2021},
  publisher={IEEE}
}

@inproceedings{glover2024edopt,
  title={EDOPT: Event-camera 6-DoF Dynamic Object Pose Tracking},
  author={Glover, Arren and Gava, Luna and Li, Zhichao and Bartolozzi, Chiara},
  booktitle={2024 IEEE International Conference on Robotics and Automation (ICRA)},
  pages={18200--18206},
  year={2024},
  organization={IEEE}
}

@article{benosman2012asynchronous,
  title={Asynchronous frameless event-based optical flow},
  author={Benosman, Ryad and Ieng, Sio-Hoi and Clercq, Charles and Bartolozzi, Chiara and Srinivasan, Mandyam},
  journal={Neural Networks},
  volume={27},
  pages={32--37},
  year={2012},
  publisher={Elsevier}
}

@inproceedings{goyal2023moveenet,
  title={MoveEnet: Online high-frequency human pose estimation with an event camera},
  author={Goyal, Gaurvi and Di Pietro, Franco and Carissimi, Nicolo and Glover, Arren and Bartolozzi, Chiara},
  booktitle={Proceedings of the IEEE/CVF Conference on Computer Vision and Pattern Recognition},
  pages={4024--4033},
  year={2023}
}

@article{gallego2020event,
  title={Event-based vision: A survey},
  author={Gallego, Guillermo and Delbr{\"u}ck, Tobi and Orchard, Garrick and Bartolozzi, Chiara and Taba, Brian and Censi, Andrea and Leutenegger, Stefan and Davison, Andrew J and Conradt, J{\"o}rg and Daniilidis, Kostas and others},
  journal={IEEE transactions on pattern analysis and machine intelligence},
  volume={44},
  number={1},
  pages={154--180},
  year={2020},
  publisher={IEEE}
}

@article{valerdi2023insights,
  title={Insights into batch selection for event-camera motion estimation},
  author={Valerdi, Juan L and Bartolozzi, Chiara and Glover, Arren},
  journal={Sensors},
  volume={23},
  number={7},
  pages={3699},
  year={2023},
  publisher={MDPI}
}

@article{shiba2022event,
  title={Event collapse in contrast maximization frameworks},
  author={Shiba, Shintaro and Aoki, Yoshimitsu and Gallego, Guillermo},
  journal={Sensors},
  volume={22},
  number={14},
  pages={5190},
  year={2022},
  publisher={MDPI}
}

@inproceedings{vasco2016fast,
  title={Fast event-based Harris corner detection exploiting the advantages of event-driven cameras},
  author={Vasco, Valentina and Glover, Arren and Bartolozzi, Chiara},
  booktitle={2016 IEEE/RSJ international conference on intelligent robots and systems (IROS)},
  pages={4144--4149},
  year={2016},
  organization={IEEE}
}

@inproceedings{mitrokhin2018event,
  title={Event-based moving object detection and tracking},
  author={Mitrokhin, Anton and Ferm{\"u}ller, Cornelia and Parameshwara, Chethan and Aloimonos, Yiannis},
  booktitle={2018 IEEE/RSJ International Conference on Intelligent Robots and Systems (IROS)},
  pages={1--9},
  year={2018},
  organization={IEEE}
}

@article{chen2022ecsnet,
  title={Ecsnet: Spatio-temporal feature learning for event camera},
  author={Chen, Zhiwen and Wu, Jinjian and Hou, Junhui and Li, Leida and Dong, Weisheng and Shi, Guangming},
  journal={IEEE Transactions on Circuits and Systems for Video Technology},
  volume={33},
  number={2},
  pages={701--712},
  year={2022},
  publisher={IEEE}
}

@inproceedings{scarpellini2021lifting,
  title={Lifting monocular events to 3d human poses},
  author={Scarpellini, Gianluca and Morerio, Pietro and Del Bue, Alessio},
  booktitle={Proceedings of the IEEE/CVF Conference on Computer Vision and Pattern Recognition},
  pages={1358--1368},
  year={2021}
}

@article{baldwin2022time,
  title={Time-ordered recent event (tore) volumes for event cameras},
  author={Baldwin, R Wes and Liu, Ruixu and Almatrafi, Mohammed and Asari, Vijayan and Hirakawa, Keigo},
  journal={IEEE Transactions on Pattern Analysis and Machine Intelligence},
  volume={45},
  number={2},
  pages={2519--2532},
  year={2022},
  publisher={IEEE}
}

@inproceedings{stoffregen2019event,
  title={Event-based motion segmentation by motion compensation},
  author={Stoffregen, Timo and Gallego, Guillermo and Drummond, Tom and Kleeman, Lindsay and Scaramuzza, Davide},
  booktitle={Proceedings of the IEEE/CVF International Conference on Computer Vision},
  pages={7244--7253},
  year={2019}
}

@inproceedings{sironi2018hats,
  title={HATS: Histograms of averaged time surfaces for robust event-based object classification},
  author={Sironi, Amos and Brambilla, Manuele and Bourdis, Nicolas and Lagorce, Xavier and Benosman, Ryad},
  booktitle={Proceedings of the IEEE conference on computer vision and pattern recognition},
  pages={1731--1740},
  year={2018}
}

@article{lagorce2016hots,
  title={Hots: a hierarchy of event-based time-surfaces for pattern recognition},
  author={Lagorce, Xavier and Orchard, Garrick and Galluppi, Francesco and Shi, Bertram E and Benosman, Ryad B},
  journal={IEEE transactions on pattern analysis and machine intelligence},
  volume={39},
  number={7},
  pages={1346--1359},
  year={2016},
  publisher={IEEE}
}

@inproceedings{zhu2018ev,
  title={EV-FlowNet: Self-Supervised Optical Flow Estimation for Event-based Cameras},
  author={Zhu, Alex Zihao and Yuan, Liangzhe},
  booktitle={Robotics: Science and Systems},
  year={2018}
}

@article{mueggler2017event,
  title={The event-camera dataset and simulator: Event-based data for pose estimation, visual odometry, and SLAM},
  author={Mueggler, Elias and Rebecq, Henri and Gallego, Guillermo and Delbruck, Tobi and Scaramuzza, Davide},
  journal={The International Journal of Robotics Research},
  volume={36},
  number={2},
  pages={142--149},
  year={2017},
  publisher={SAGE Publications Sage UK: London, England}
}

@inproceedings{mitrokhin2019ev,
  title={EV-IMO: Motion segmentation dataset and learning pipeline for event cameras},
  author={Mitrokhin, Anton and Ye, Chengxi and Ferm{\"u}ller, Cornelia and Aloimonos, Yiannis and Delbruck, Tobi},
  booktitle={2019 IEEE/RSJ International Conference on Intelligent Robots and Systems (IROS)},
  pages={6105--6112},
  year={2019},
  organization={IEEE}
}

@inproceedings{iacono2018towards,
  title={Towards event-driven object detection with off-the-shelf deep learning},
  author={Iacono, Massimiliano and Weber, Stefan and Glover, Arren and Bartolozzi, Chiara},
  booktitle={2018 IEEE/RSJ International Conference on Intelligent Robots and Systems (IROS)},
  pages={1--9},
  year={2018},
  organization={IEEE}
}

@article{niu2025esvo2,
  title={Esvo2: Direct visual-inertial odometry with stereo event cameras},
  author={Niu, Junkai and Zhong, Sheng and Lu, Xiuyuan and Shen, Shaojie and Gallego, Guillermo and Zhou, Yi},
  journal={IEEE Transactions on Robotics},
  year={2025},
  publisher={IEEE}
}

@article{hirschmuller2007stereo,
  title={Stereo processing by semiglobal matching and mutual information},
  author={Hirschmuller, Heiko},
  journal={IEEE Transactions on pattern analysis and machine intelligence},
  volume={30},
  number={2},
  pages={328--341},
  year={2007},
  publisher={IEEE}
}

@article{zhu2018multivehicle,
  title={The multivehicle stereo event camera dataset: An event camera dataset for 3D perception},
  author={Zhu, Alex Zihao and Thakur, Dinesh and {\"O}zaslan, Tolga and Pfrommer, Bernd and Kumar, Vijay and Daniilidis, Kostas},
  journal={IEEE Robotics and Automation Letters},
  volume={3},
  number={3},
  pages={2032--2039},
  year={2018},
  publisher={IEEE}
}

@article{Grompone2012,
    title   = {{{LSD}: a Line Segment Detector}},
    author  = {Grompone von Gioi, Rafael and Jakubowicz, Jérémie and Morel, Jean-Michel and Randall, Gregory},
    journal = {{Image Processing On Line}},
    volume  = {2},
    pages   = {35--55},
    year    = {2012},
}

@INPROCEEDINGS{Huang2018,
  author={Huang, Kun and Wang, Yifan and Zhou, Zihan and Ding, Tianjiao and Gao, Shenghua and Ma, Yi},
  booktitle={2018 IEEE/CVF Conference on Computer Vision and Pattern Recognition}, 
  title={Learning to Parse Wireframes in Images of Man-Made Environments}, 
  year={2018},
  volume={},
  number={},
  pages={626-635},
  doi={10.1109/CVPR.2018.00072}}

@INPROCEEDINGS{Zhou2019,
  author={Zhou, Yichao and Qi, Haozhi and Ma, Yi},
  booktitle={2019 IEEE/CVF International Conference on Computer Vision (ICCV)}, 
  title={End-to-End Wireframe Parsing}, 
  year={2019},
  volume={},
  number={},
  pages={962-971},
  doi={10.1109/ICCV.2019.00105}}

@inproceedings{zhu2019unsupervised,
  title={Unsupervised event-based learning of optical flow, depth, and egomotion},
  author={Zhu, Alex Zihao and Yuan, Liangzhe and Chaney, Kenneth and Daniilidis, Kostas},
  booktitle={Proceedings of the IEEE/CVF conference on computer vision and pattern recognition},
  pages={989--997},
  year={2019}
}
